\documentclass[12pt, oneside]{article}   	
\usepackage{amssymb}
\usepackage{amsmath}
\usepackage{adjustbox}
\usepackage{array}
\usepackage{authblk}
\usepackage{bm}
\usepackage{booktabs}
\usepackage{breakcites}
\usepackage{csquotes}
\usepackage{comment}
\usepackage{enumitem}
\usepackage{geometry}
\usepackage{graphicx}
\usepackage{hyperref}
\usepackage{setspace}
\usepackage{xcolor}
\usepackage{url}

\usepackage{lineno}

\title{Accuracy and Fairness of Facial Recognition Technology in Low-Quality Police Images: An Experiment With Synthetic Faces}

\author[1,2]{Maria Cuellar \footnote{Corresponding author: Maria Cuellar, mcuellar@sas.upenn.edu.}}
 
\author[1]{Hon Kiu (James) To}

\author[3]{Arush Mehrotra}

\affil[1]{Department of Criminology, University of Pennsylvania, 3718 Locust Walk, Philadelphia, PA, 19104}

\affil[2]{Department of Statistics and Data Science, Wharton School, University of Pennsylvania, Walnut Street, Philadelphia, PA 19104}

\affil[3]{Computer Science, School of Engineering \& Applied Science, University of Pennsylvania, 220 S 33rd St, Philadelphia, PA 19104}

\begin{document}
\maketitle

\begin{abstract}

Facial recognition technology (FRT) is increasingly used in criminal investigations, yet most evaluations of its accuracy rely on high-quality images, unlike those often encountered by law enforcement. This study examines how five common forms of image degradation--contrast, brightness, motion blur, pose shift, and resolution--affect FRT accuracy and fairness across demographic groups. Using synthetic faces generated by StyleGAN3 and labeled with FairFace, we simulate degraded images and evaluate performance using Deepface with ArcFace loss in 1:n identification tasks. We perform an experiment and find that false positive rates peak near baseline image quality, while false negatives increase as degradation intensifies--especially with blur and low resolution. Error rates are consistently higher for women and Black individuals, with Black females most affected. These disparities raise concerns about fairness and reliability when FRT is used in real-world investigative contexts. Nevertheless, even under the most challenging conditions and for the most affected subgroups, FRT accuracy remains substantially higher than that of many traditional forensic methods. This suggests that, if appropriately validated and regulated, FRT should be considered a valuable investigative tool. However, algorithmic accuracy alone is not sufficient: we must also evaluate how FRT is used in practice, including user-driven data manipulation. Such cases underscore the need for transparency and oversight in FRT deployment to ensure both fairness and forensic validity.

\end{abstract}

\noindent%

\section{Introduction}\label{sec:introduction}

Law enforcement agencies increasingly rely on facial recognition technology (FRT), a method that uses artificial intelligence (AI) to identify suspects from still images. Tools like Clearview AI, Amazon Rekognition, and Oosto, are used by police and ICE, providing access to over 50 billion identified images scraped from public websites, DMV records, border crossings, and more. FRT has become commonplace in criminal investigations, with at least 3,750 state and local law enforcement agencies and 20 federal agencies currently reporting using FRT, and many agencies have plans to expand their use of FRT in recent years (\cite{Garvie2022}). 

\begin{figure}[ht]
   \centering
   \includegraphics[width=.8\textwidth]{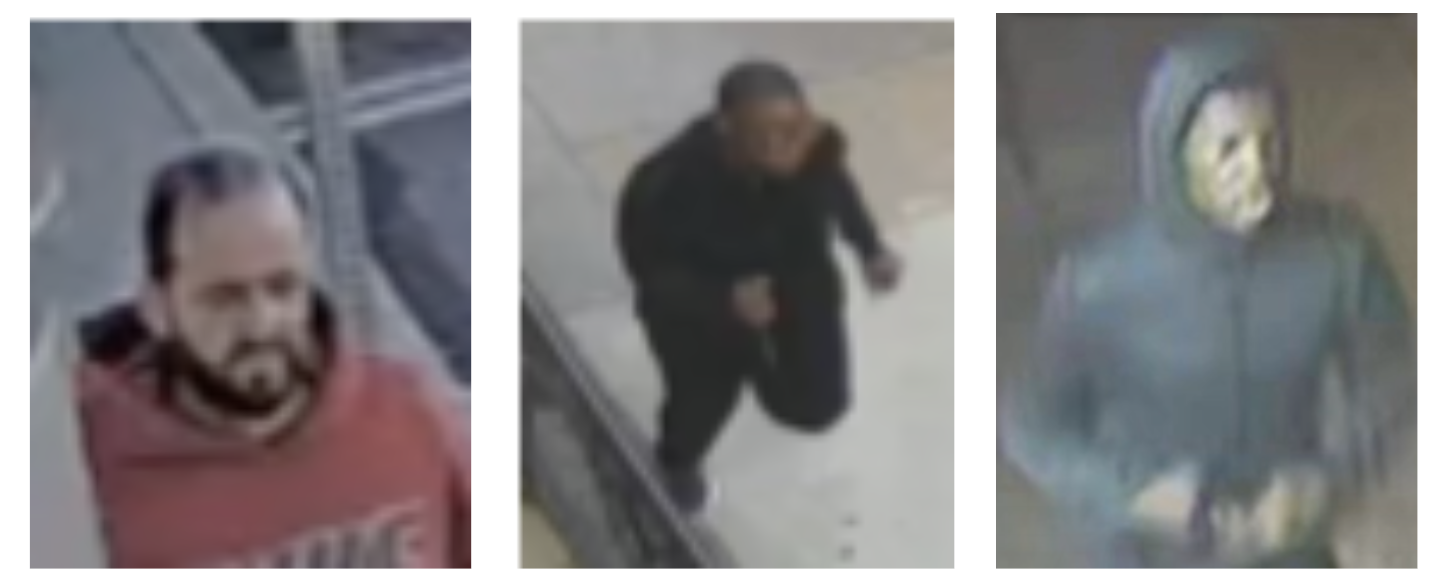}
   \caption{Images released by the Philadelphia Police Department in public appeals for help identifying individuals suspected of crimes often exhibit low contrast and brightness, significant blur, non-frontal poses, and low resolution. Images from the official YouTube channel of the Philadelphia Police Department, \url{https://www.youtube.com/user/PhiladelphiaPolice}, obtained 05/05/2025.}
   \label{fig:ppdimages}
\end{figure}

Although FRT is increasingly used, and often promoted as highly accurate (\cite{clearviewaiaccuracy}), the accuracy of FRTs has been tested primarily in settings with high-quality images (\cite{nistvendortest}). It is possible that the algorithms work less well in low-quality images such as those often used by law enforcement (see some examples of these in Figure \ref{fig:ppdimages}). But, the promotion of FRTs being so accurate can lead to law enforcement being misled. For example, a media investigation revealed that some law enforcement officers treat FRT outputs as definitive rather than investigative leads, for example, referring to an unverified match as `100\% match' (\cite{wapostinvestigation}). However, the Innocence Project reports that there are ``at least seven confirmed cases of misidentification due to facial recognition technology, six of which involve Black people who have been wrongfully accused.'' (\cite{ipfrt}). 

But how often are FRTs making errors? And is this happening equally throughout demographic groups? Without proper validation, using FRT in practice risks leading to a wrongful conviction or wrongful acquittal, and this could be worse for individuals of certain demographic characteristics. Law enforcement officers and other stakeholders might trust the FRT results more or less than they should.

There is some research on how image quality influences FRT accuracy (\cite{Zhang_etal2018}, \cite{Liu_etal2008}, \cite{Punnappurath_etal2015}, \cite{Bacci_etal2024}, \cite{terhorst2023pixel}), but it is unclear how this varies across various levels and types of image quality. A notable resource is the vendor test performed by the National Institute on Standards and Technology, which tested over 100 commonly used FRT algorithms and reported their performance in 1:1 and 1:n tasks, across various factors, including race \cite{nistvendortest}. Several factors can affect image quality, including but not limited to, object motion (\cite{Zhang_etal2018}, \cite{Liu_etal2008}, \cite{Punnappurath_etal2015}), camera specifications and placement (\cite{Bacci_etal2024}, \cite{Zhang_etal2018}, \cite{Liu_etal2008}), illumination (\cite{Bacci_etal2024}, \cite{Punnappurath_etal2015}), pose (\cite{Punnappurath_etal2015}), and facial expression (\cite{Punnappurath_etal2015}). 

Differential accuracy of FRTs by gender and race (i.e., what we call a lack of fairness) is another critical concern. Studies show that false positive rates are higher for Black individuals, women, and certain age groups (\cite{Grother_etal2019_Part3}, \cite{Cavazos_etal2020}, \cite{Liang_etal2023}, \cite{bestrowden2018learning}), a fact that risks the constitutionally protected equal protection under the law. Different methods have been proposed to measure fairness (\cite{howard2019effect}, \cite{howard2020quantifying}, \cite{howard2022disparate}, \cite{howard2022evaluating}). However, a comprehensive evaluation of FRT accuracy by race and gender in low-quality images is lacking.

Courts have begun to require disclosure when a lead was generated using FRT, and it is likely that FRT results will increasingly be introduced in evidentiary contexts. As of early 2025, 15 states had enacted laws regulating FRT use in policing (\cite{minnesotareformer}), and courts are beginning to require disclosure when FRT is used. As such, courts' admissibility standards for scientific evidence offer a valuable framework for evaluating the validity of these tools. Legal scholars have raised concerns about transparency, proprietary algorithms, and contextual bias (\cite{jackson2019challenging}, \cite{njvsarteaga}, \cite{aclufacial}, \cite{haddad2020confronting}, \cite{celentino2015face}). As FRT moves closer to evidentiary use, it must meet the same scientific standards courts apply to other forensic methods (\cite{PCAST2016}).

Our approach draws on lessons learned from problems in traditional forensic science disciplines, specifically, their lack of demonstrated reliability (accuracy, repeatability, and reproducibility) (\cite{PCAST2016}). It applies the corresponding solutions proposed in the forensic science reform literature (\cite{stern2019reliability}) to the study of facial recognition technologies (FRTs). While FRT is not typically categorized as a forensic science and is not yet widely introduced as evidence in court, its use in criminal investigations plays a similar role: identifying suspects. And, so far, it seems to be more successful than other forensic disciplines, as shown in Table \ref{tab:error_rates}.

\begin{table}[ht!]
\centering
\begin{tabular}{lcc}
\toprule
\textbf{Forensic Discipline} & \textbf{False Positive Rate} & \textbf{False Negative Rate} \\
\midrule
Facial recognition (FRT) & 0.0001\% & 0.06\% \\
Fingerprint comparison              & 0.1\%    & 7.5\%  \\
Firearm comparison                 & 1.01\%   & 0.37\% \\
\bottomrule
\end{tabular}
\caption{Example error rates for FRT (\cite{Grother_etal2019_Part1}) compared to fingerprint (\cite{ulery2011}) and firearm (\cite{amesi}) comparisons, as they are currently practiced. When applied to high-quality images, facial recognition technology (FRT) outperforms many traditional forensic disciplines in terms of accuracy. However, it is important to note that the reported error rates come from studies that have faced methodological criticism (\cite{cuellarfirearms}).}
\label{tab:error_rates}
\end{table}

In this study, we evaluate FRT performance by isolating the effects of image quality (contrast, brightness, motion blur, pose, and resolution) on accuracy and fairness. Using simulated data (ironically) gives us a more realistic picture of the accuracy and fairness of the technology. Synthetic faces allow for gradual degradation of the images for the generation of error-rate curves, conditional on race and gender (\cite{melzi2024frcsyn}, \cite{du2024impact}, \cite{rahimi2024synthetic}). This is a rapidly changing area of research, both because algorithms are being updated often and because the use by law enforcement and the legal restrictions for its use are also changing. Providing a more comprehensive evaluation of FRT, including measures of accuracy and fairness, can help law enforcement gauge the actual reliability of algorithms and devise a more proper use of FRT results, such as treating them as merely investigative leads or probable cause.

\section{Materials}

\subsection{Synthetic face generation}

\begin{figure}[ht]
    \centering
    \includegraphics[width=.5\linewidth]{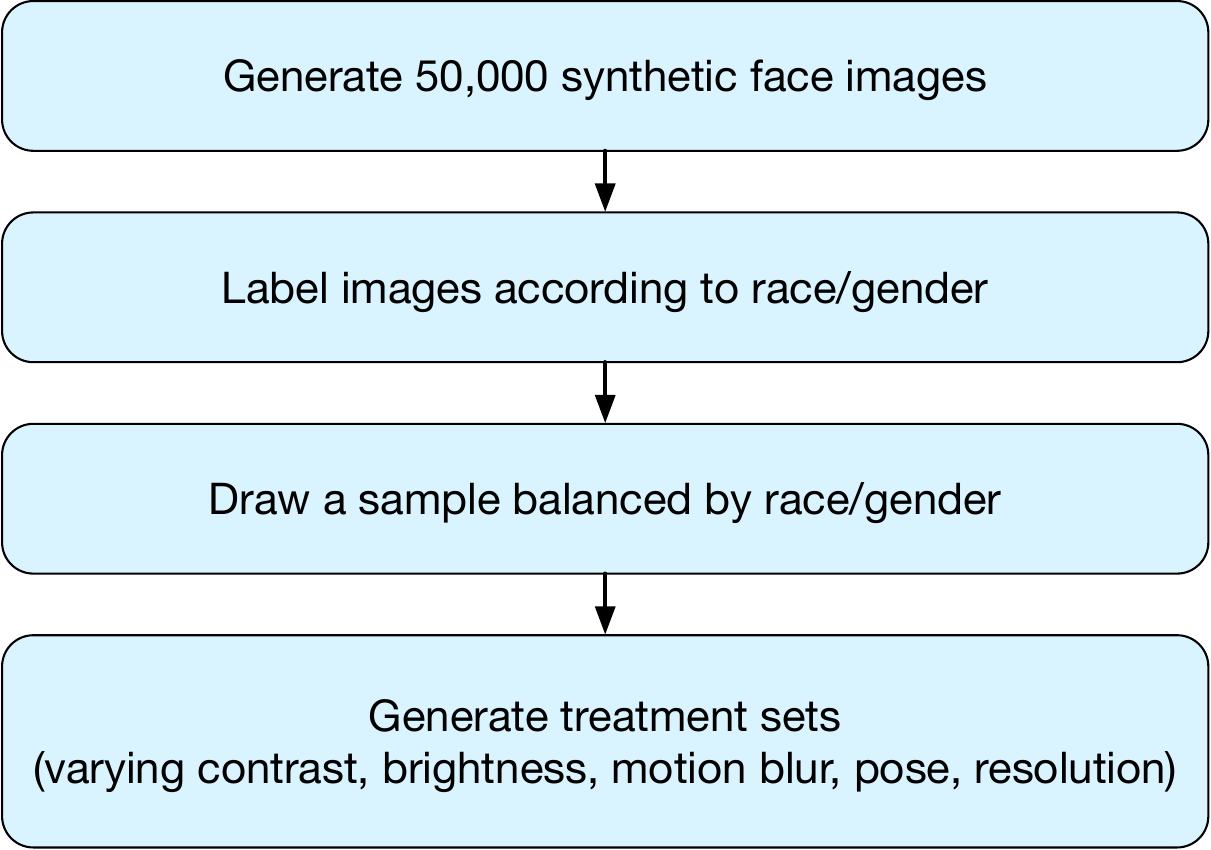}
    \caption{Data generation procedure. We generate a large population of synthetic faces, and then take a random sample that is balanced by race and gender. We then use this to generate treatment and control groups.}
    \label{fig:data-generation}
\end{figure}

To assess the impact of image quality on FRT performance, we generate a dataset of synthetic faces, following the process outlined in Figure \ref{fig:data-generation}. Synthetic data allows for precise experimental manipulation of image quality while eliminating confounding variables associated with real-world datasets. It is also a way to rotate the face's pose experimentally for a large number of faces at once. The dataset consists of a ``control'' set, which includes high-quality images, and several ``treatment'' sets where images have been degraded in specific ways to simulate real-world conditions. The treatment images include variations in contrast, brightness, motion blur, pose, and resolution, allowing for a detailed analysis of how these factors influence FRT accuracy.

A state-of-the-art facial recognition algorithm was used to compare synthetic images against a database, measuring both false-positive and false-negative rates. The false positives are instances for which the algorithm incorrectly identifies an individual as a match, while the false negatives are instances for which the algorithm incorrectly rejects a match. Error rates were analyzed across different racial and gender groups to determine whether FRT exhibits systematic biases that could affect certain populations.

We first generate base synthetic faces using StyleGAN3 (\cite{Karras_etal2021}). StyleGAN3 is a popular face generation model that is capable of generating high-quality, photo-realistic faces (\cite{Alaluf_etal2022,Grissom_etal2024}). Users can freely choose their desired pre-trained network for generation, as well as a truncation parameter $0 \leq \psi \leq 1$ (\cite{Karras_etal2019,Mundra_etal2023}). The pre-trained network is the training sample (i.e., the set of faces used to train the network and let it learn the features and patterns of the faces) for face generation, and the truncation parameter $\psi$ controls the degree of weighting away from the ``average'' face of the pre-trained network.
In particular, choosing $\psi = 0$ always returns the ``average'' face of the pre-trained network. The truncation parameter $\psi$ also plays a role in facial variety. A smaller $\psi$ tends to reduce the facial variety of the generated faces (see, e.g., \cite{Karras_etal2019, Mundra_etal2023, Maluleke_etal2022}). Some common choices of $\psi$ are 0.5 and 0.7. 

\begin{figure}[ht]
    \centering
    \includegraphics[width=.6\linewidth]{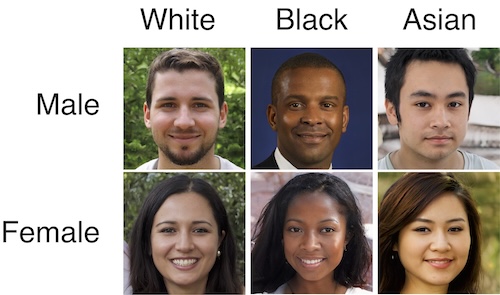}
    \caption{Examples of synthetic faces generated with StyleGAN3 and labeled with FairFace.}
    \label{fig:races}
\end{figure}

For this study, we select the pre-trained network based on the Flickr-Faces-HQ (FFHQ) dataset, which consists of 70,000 high-quality PNG images at 1024$\times$1024 resolution and contains considerable variation in terms of age, ethnicity and image background. It also includes photos with accessories such as eyeglasses, sunglasses, hats, etc. The images were crawled from Flickr, thus inheriting all the biases of that website, and automatically aligned and cropped using dlib. A truncation parameter of $\psi = 0.7$ was used (\cite{Karras_etal2019}). Figure \ref{fig:races} shows some examples of faces generated with StyleGAN3 based on the FFHQ network and a truncation parameter of $\psi = 0.7$, and classified using FairFace in terms of gender and three races. Utilizing the StyleGAN3 algorithm, we generate an initial database consisting of 50,000 faces.

\subsection{Sampling scheme}

From the pool of 50,000 generated faces, we randomly sample 167 as the ``control'' database. The sampling is performed in a way that the gender and racial composition of the database mimics the actual composition of the United States as closely as possible so that we have a control database that is representative of the United States population. The number of control faces, 167, is selected based on the \texttt{MRMCsamplesize} package (\cite{Robert_etal2023}), based on the \texttt{sampleSize\_Standalone} function using an expected sensitivity of 0.99 with half-width 0.025 for confidence intervals under 90\% assumed power. We randomly sample 167 ``treated'' faces for comparison against the control database. In particular, the sampling is done so that 84 faces do not have their corresponding control faces in the database (target-absent), and the remaining 83 faces have their corresponding control faces in the database (target-present). Our sampling process gives a Monte Carlo approximation to a 95\% confidence interval. This is an empirical confidence interval constructed via repeated sampling from the population, a process that we can do here because we have a large sample of synthetic data.

\subsection{Race and gender sample distribution}

We create a balanced sample of demographic features of the generated faces. To do this, we first assess the demographic features of our generated faces using FairFace, a facial attribute analysis tool. FairFace predicts the demographic labels, including races, age groups, and genders (\cite{KarkkainenJoo2021}). The FairFace algorithm has been shown to have high predictive performances across race, gender, and age groups (\cite{KarkkainenJoo2021}), and is used for labeling faces  (\cite{Melzi_etal2023}).

Regarding age, we remove all the children from our sample, who were labeled by FairFace as being under age 10. The age 10 is the minimum age of juvenile adjudication in 15 states and territories (\cite{NGA2021}) or the minimum age of criminal responsibility in other places, such as Hong Kong (\cite{JuvenileOffendersOrdinance2022}), England and Wales (\cite{GOVUK_AgeCriminalResponsibility_2025}), and South Australia (\cite{SAGovMACR2025}). Regarding race, we adopt four races provided by FairFace (White, Asian, Black, and Indian), but we extract only Black, White, and Asian for simplicity. The original ``Asian'' and ``Indian'' are combined into one single ``Asian'' group. A more nuanced treatment of race (e.g., include more races and account for ethnicity) is important and left for future research. The faces are generated from a pre-trained network that utilizes the FFHQ dataset as training samples, so it inherits the existing distribution of the training dataset and reflects in the generated faces. 

Our original sample is 73\% white 17\% Asian, 5\% Black, and 5\% Indian, while the 2020 US Census (\cite{USCensus2020}) states that the US population is 57.8\% white (non-Hispanic), 5.9\% Asian, 12.1\% Black, 0.7\% American Indian and Alaska Native alone (non-Hispanic). Percentages do not sum to 100\% because we select specific races to study in our experiment. The gender distribution of the generated faces is well-balanced, and the distribution also closely resembles that of the United States, which has 50.4\% females. We do not study age distributions here. We use a simple random sampling scheme to balance our sample so it matches the 2020 US Census racial distribution. Table \ref{tab:summarystats} shows the demographic distributions of our synthetic face control sample.

\begin{table}[ht]
    \centering
    \caption{Summary statistics of the sample of synthetic faces we used for our experiment. This is the overall summary by gender and race after dropping children, combining the selected races, and dropping problematic images that raise exceptions when editing their poses.}
        \label{tab:summarystats}
    \begin{tabular}{lcc}
        \toprule
        & \textbf{Count} & \textbf{Percentage} \\
        \midrule
        \textbf{Gender} \\
        \hspace{5mm} Female & 22,291 & 49.937\% \\
        \hspace{5mm} Male & 22,247 & 50.063\%  \\
        \midrule
        \textbf{Race} \\
        \hspace{5mm} White & 32,425 & 72.967\% \\
        \hspace{5mm} Black & 2,298 & 5.171\%  \\
        \hspace{5mm} Asian & 9,715 & 21.862\%  \\
        \midrule
        \textbf{Total cases} & 44,438 & 100\%  \\
        \bottomrule
    \end{tabular}
\end{table}

\subsection{Image degradation factors}

\begin{figure}[t]
    \centering
    \includegraphics[width=\linewidth]{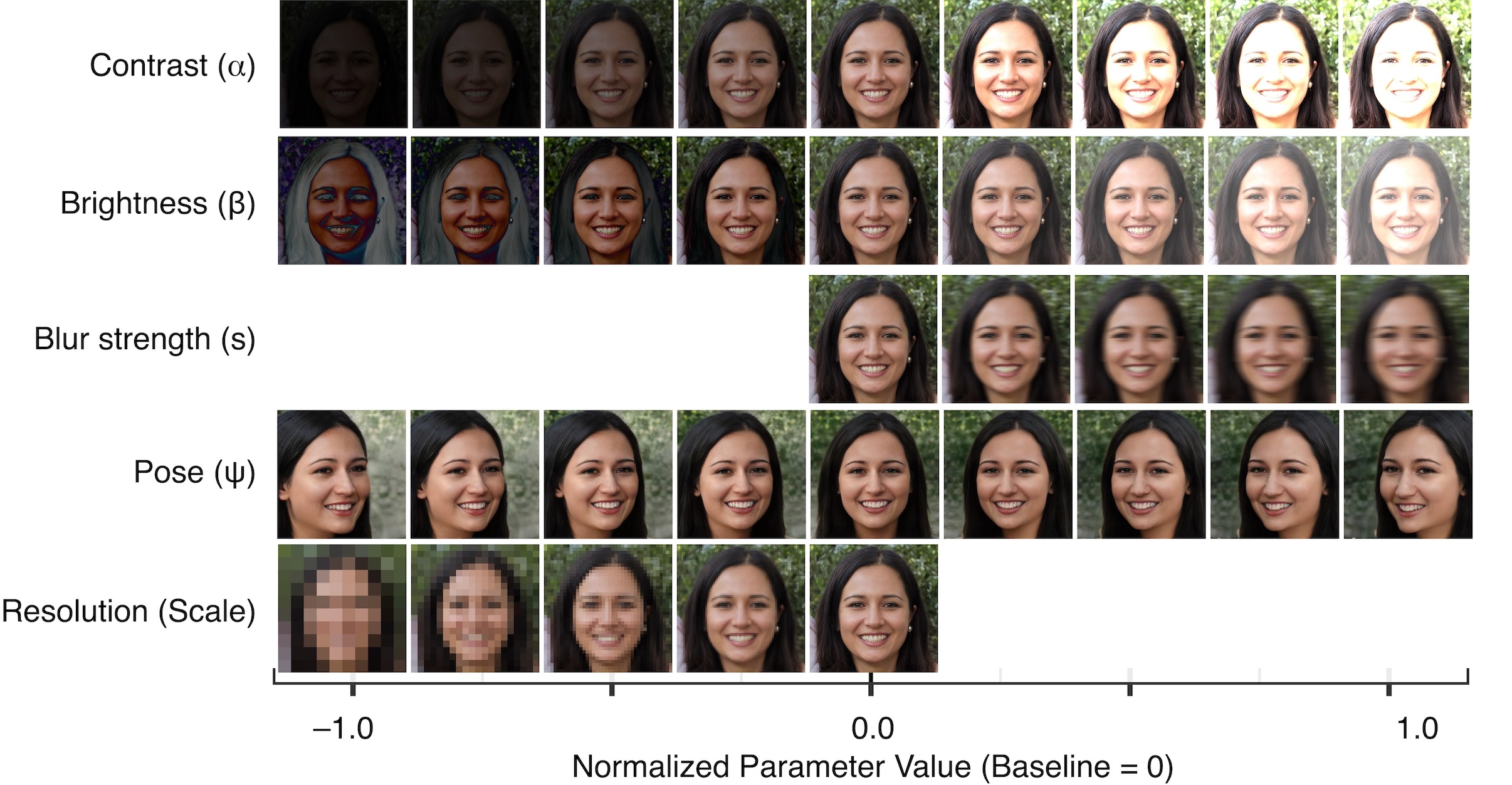}
    \caption{Facial images that have been modified according to parameters (brightness, contrast, motion blur, resolution, and pose). The image in the center is the baseline, control image.}
    \label{fig:treatments}
\end{figure}

The next step is editing the generated faces to alter the image qualities at various levels. Figure \ref{fig:treatments} shows example images that have been degraded. We normalize the parameter values for all degradation factors (from -1 to 1, with 0 being the baseline, undegraded image) to do inter-factor comparisons. For this paper, we consider five factors: contrast, brightness, motion blur, pose, and resolution. We use Python's \texttt{cv2} module (\cite{opencv_library}) for editing the brightness, contrast, and resolution, Python's \texttt{blurgenerator} (\cite{lee_blur_generator}) module for generating motion blurs, and the InterFaceGAN module (\cite{Shen_etal2020}), implemented through the codes provided by \cite{Alaluf_etal2022} for editing the pose. Note that motion blur and resolution are only present above and below zero, respectively, because of the nature of the image degradation (we do not de-blur the original image further, and we do not try to get a better resolution than the original image's). Appendix A provides more details about the parameters we used to generate the treatment groups.

\section{Method}

\begin{figure}[t]
    \centering
    \includegraphics[width=.5\linewidth]{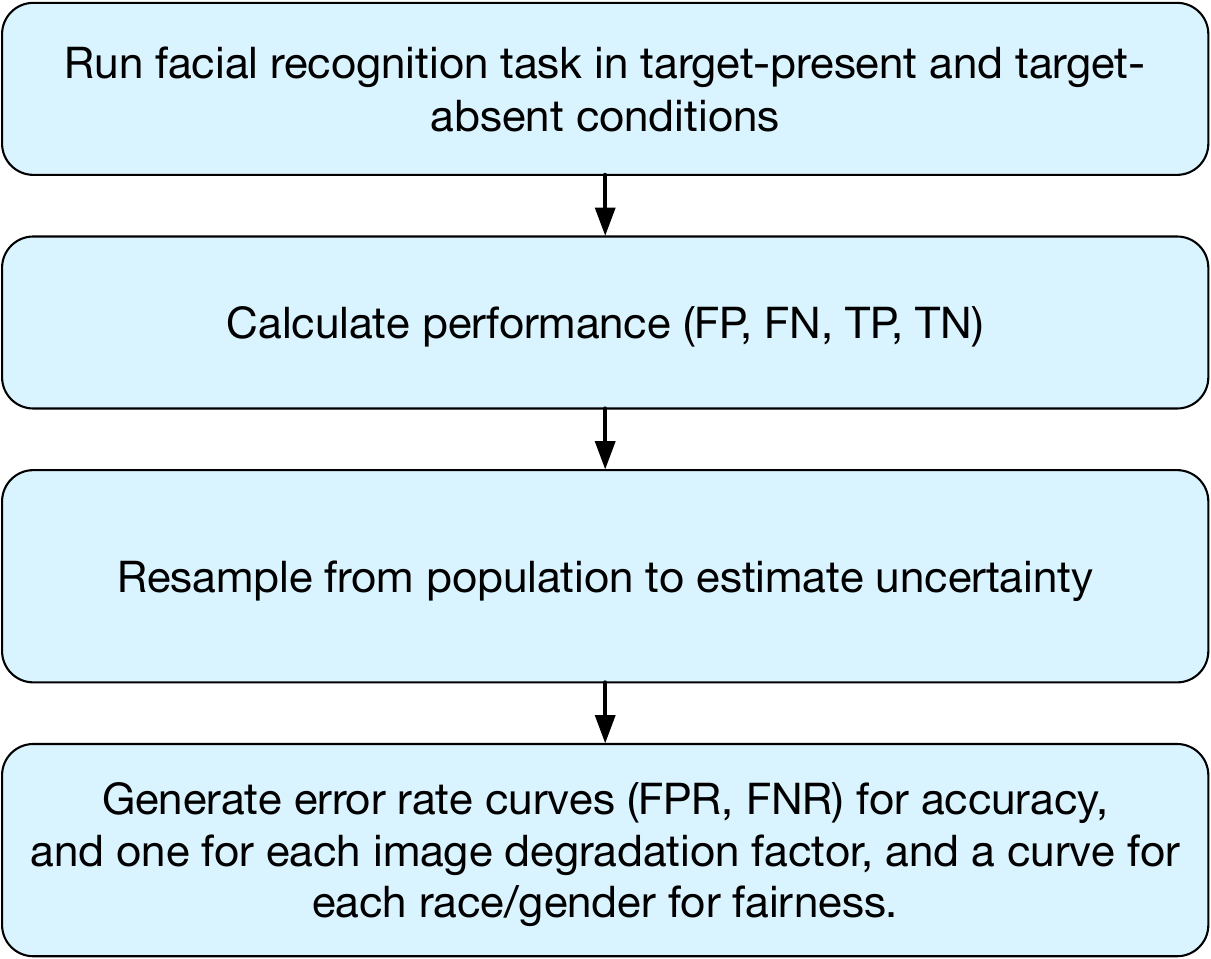}
    \caption{Our method to generate the conditional error rate curves.}
    \label{fig:analysis}
\end{figure}

\subsection{FRT algorithm}

Our experimental design is outlined in Figure \ref{fig:analysis}. As our FRT algorithm to evaluate, we selected Deepface, a lightweight hybrid FRT framework that wraps the state-of-the-art facial recognition models (\cite{SerengilOzpinar2020}). In particular, we chose the RetinaFace model (\cite{Deng_etal2020}) for face detection and the ArcFace model (\cite{Deng_etal2019}) for facial recognition, both are developed in the InsightFace project. ArcFace is recognized for its high accuracy and efficiency, and has been evaluated in NIST's FRVT, where it stands out among open-source algorithms (\cite{Bolme_etal2020,nist_frvt_visa}). The continued active development and strong empirical results ensure these models remain relevant and adaptable for future facial recognition challenges. To ensure that all ``treated'' images, including those with severe quality degradation, could be processed by the face recognition system, we set ``enforce\_detection = False'' in DeepFace. This setting allows the system to compute embeddings even when automatic face detection fails, thereby preventing exceptions and enabling a comprehensive evaluation of performance under challenging conditions. While this approach may reduce accuracy, it reflects the need to assess system robustness with realistic, low-quality images.

\subsection{Error rates of 1:n identification task}

We perform a total of $2^{8} = 256$ replications for the simulation studies. For every treated faces, we perform a 1-to-167 facial recognition task using Deepface with the ArcFace loss (\cite{Deng_etal2019}). For the facial recognition outputs resulting from each of the 1-to-167 comparisons, we count the number of true positives, true negatives, false positives, and false negatives as follows: For the target-absent group, a match is counted as a false positive, and a non-match is counted as a true negative. For the target-present group, a correct match is counted as a true positive, an incorrect match is counted as a false positive. Similarly, a correct non-match is counted as a true negative, and an incorrect non-match is counted as a false negative.

After performing the facial recognition tasks for all replications following the above steps, we compute the number of false positives, false negatives, true positives, and true negatives for each replication. For each replication task, we calculate the false positive rate (FPR) with its usual definition,
\begin{equation}
\text{FPR} = \frac{\text{FP}}{\text{FP+TN}},
\end{equation}
and the false negative rate as,
\begin{equation}
\text{FNR} = \frac{\text{FN}}{\text{FN+TP}}.
\end{equation}

The facial recognition literature (\cite{nistvendortest}) distinguishes between error rates in ``verification'' tasks (1:1 image comparisons) and ``identification'' tasks (1:n search, when a system tries to identify a person from a gallery, e.g., police using a photo to search a database). Instead of a FNR, it uses FNIR (False Negative Identification Rate), the probability that the system fails to identify a person who is in the database. And instead of a FPR, it uses a FPIR (False Positive Identification Rate), the probability that the system identifies a person who is in the database. 

The notation FNIR($N$, $R$, $T$) includes three key parameters. $N$ refers to the gallery size, or the number of identities stored in the enrollment database. $R$ denotes the number of candidates returned--often called the rank or top-$R$--meaning the system returns the top $R$ matches, and identification is considered successful if the correct identity appears among them. $T$ represents the threshold, a minimum similarity score required for a match to be accepted; this value may be fixed or varied to analyze system performance across different operating points. 

We use a fixed $N$ and $T$ in this study because this is akin to what law enforcement officers will likely do as they use FRTs. $N$ is selected with our sample size calculation and $T$ is set by the recommendation of the developers of the FRT algorithm we use. $R$ is not predetermined; instead, it is determined dynamically by the threshold $T$, that is, for each image, all gallery images with a cosine distance less than or equal to $T$ are returned as matches. This mirrors a common law enforcement practice, where officers review all candidates above a certain confidence level rather than a fixed number of top matches. Note that lower cosine distance indicates greater similarity between faces.

\subsection{Fairness}

In this study, we analyze fairness by comparing error rates across race and gender groups. However, we do not employ more advanced fairness metrics or causal frameworks such as those proposed in prior work (e.g., \cite{howard2019effect}, \cite{howard2020quantifying}, \cite{howard2022disparate}, \cite{howard2022evaluating}), which offer deeper insights into the mechanisms driving disparate performance.

\section{Results}\label{sec:results}

\subsection{Accuracy}

\begin{figure}[ht]
   \centering
   \includegraphics[width=\textwidth]{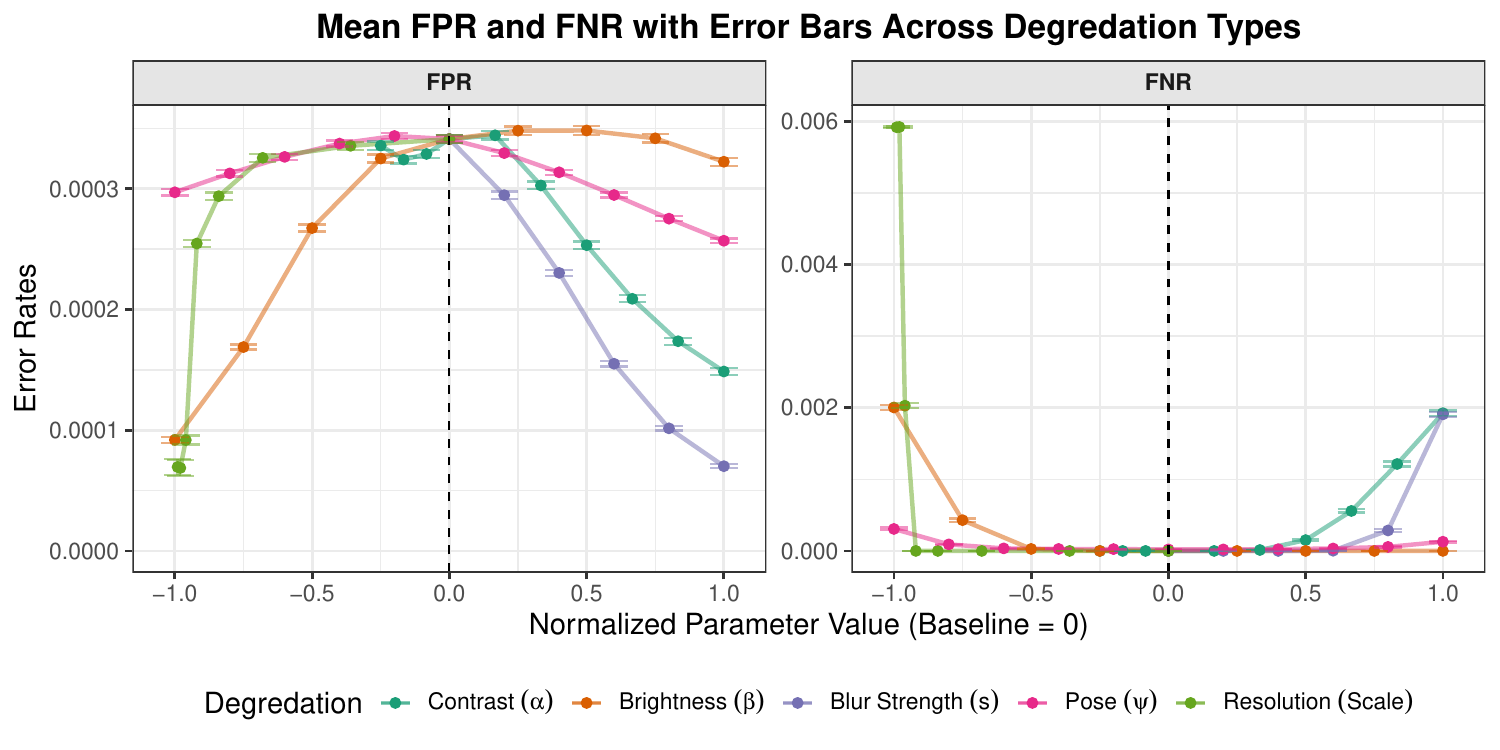} 
   \caption{Accuracy curves. False positive and false negative rates for different image degradation factors. The error bars denote the 95\% confidence intervals obtained by resampling from the large population. The vertical dashed line represents the baseline image level.}
   \label{fig:accuracy-results}
\end{figure}

Figure \ref{fig:accuracy-results} shows the results for evaluating the accuracy of the FRT algorithm. It displays the false positive rates (FPRs) and false negative rates (FNRs) of the algorithm, conditional on the image degradation factor. There is a curve for each of the factors: contrast, brightness, blur strength, pose, and resolution. The $x$-axis shows a normalized parameter value (with the baseline at 0). This represents the severity of each degradation type, normalized such that 0 corresponds to the baseline (i.e., no degradation). The axis ranges from -1.0 to 1.0. The $y$-axis shows the FPR and FNR values. Appendix \ref{sec:appendix} describes the process for normalizing these factors. 

Note that the highest false positive rates are much smaller than the highest false negative rates, by a factor of 20, indicating that false positives are rarer than false negatives.

The algorithm produces the most false positives when images are near their original, high-quality state, likely because both true and false matches yield higher similarity scores. As degradation increases, these spurious similarities diminish, leading to a drop in FPR. This inverted U-shaped pattern suggests that degradation acts as a kind of natural filter, suppressing unintended similarity between different identities.

There is high sensitivity to low contrast and high brightness. If the image contrast is low or the brightness is too high, there can be many false positives. Pose has less of an effect on changing the FPRs, and in fact has a lower FPR than other factors. 

For the false negative rate (FNR), as image quality declines, the FNR increases. Notably, blur and reduced resolution have the most pronounced effect, indicating that the algorithm is more likely to fail to recognize a true match under these conditions. In contrast, changes in brightness and contrast have a smaller impact on FNR. This raises concerns about the reliability of FRT in law enforcement settings, where low resolution and high blur images are common.

We note that pose requires a separate discussion. The reason that it can be altered gradually, at such a massive scale, and with such low cost, is because we are using synthetic data. Since many police images of suspects show faces that are not directly facing the camera, accounting for pose variation is essential for understanding error rates in real-world conditions.

\subsection{Fairness}

\begin{figure}[h!]
   \centering
   \includegraphics[width=\textwidth]{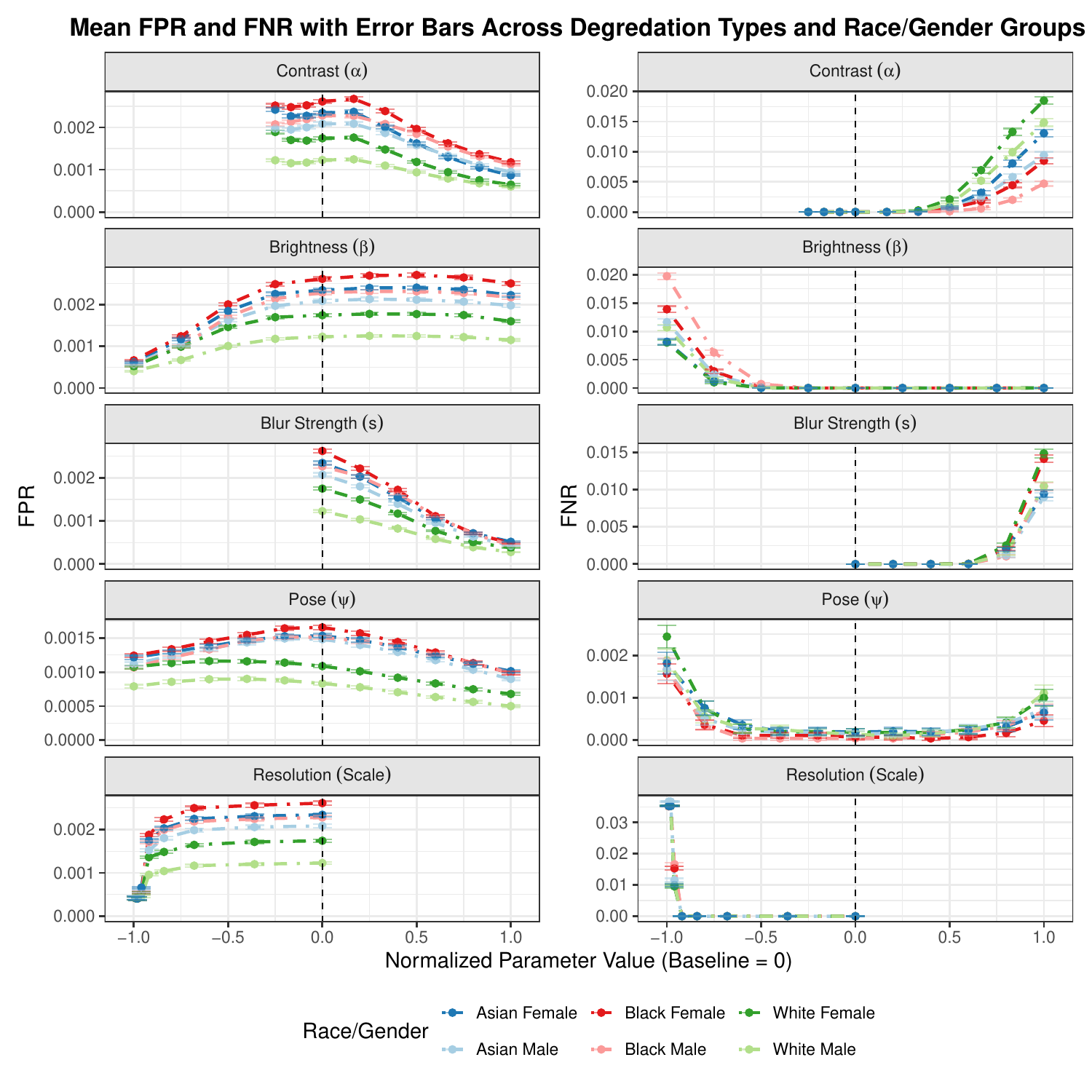} 
   \caption{Fairness curves, i.e., differential accuracy based on demographic factors: race and gender. False positive and false negative rates for different image degradation factors. The error bars denote the 95\% confidence intervals obtained by resampling from the large population. The vertical dashed line represents the baseline image level.}
   \label{fig:fairness-results}
\end{figure}

Figure \ref{fig:fairness-results} presents how the marginal conditional rates change across different demographic factors: race and gender. Across all degradation types, there are clear disparities between race/gender groups. For instance, FPR tends to be higher for Black individuals, particularly Black females, under many degradation conditions. This agrees with the literature on racial disparities in facial recognition technology (\cite{nistvendortest3}). White males often have the lowest FPR. These disparities grow less pronounced as degradation intensifies. 

These corresponding plots show how FNR varies across the same degradation types and demographic groups. The disparities here are notable: Asian and Black females generally experience higher FNRs, meaning the algorithm is more likely to miss a correct identification for them under degraded conditions. Pose and resolution degradation produce especially large increases in FNR across all groups, but the effects are unevenly distributed, with females and Black and Asian individuals disproportionately affected. In this case, the disparities grow more pronounced as degradation intensifies.

\section{Discussion}\label{sec:discussion}

Our experiment finds that facial recognition technology (FRT) performance degrades under poor image conditions, particularly with blur, pose variation, and reduced resolution, and that this degradation is not evenly distributed across demographic groups. False positive and false negative rates increase with image degradation, disproportionately affecting individuals from marginalized race and gender groups. These disparities raise important concerns about the fairness and reliability of FRT when used in real-world law enforcement contexts, where image quality is often suboptimal. While this decision provides a conservative estimate of system performance under real-world conditions, it may further reduce accuracy compared to studies that exclude such images. As a result, our findings likely represent a lower bound on expected performance, particularly in operational settings where face detection is a prerequisite for matching.

Nevertheless, it is worth noting that even under the most challenging conditions and for the most affected subgroups, the accuracy of FRT remains substantially higher than that of many traditional forensic methods. This suggests that, if appropriately validated and regulated, FRT should be considered a useful investigative tool. However, the accuracy of the algorithm alone is not sufficient: we must also assess the accuracy of actual use in practice, including how human users pre-process and manipulate the data. For example, there is a documented case in which a police officer copied facial features from high-resolution images and pasted them onto a low-quality suspect photo using computer software prior to conducting a database search \cite{Garvie2022}. This eggregious example shows that it is crucial to understand how FRT is being used in practice, and how this affects the accuracy and fairness measures.

According to our experiment, facial recognition systems are most likely to generate false positives when the image quality appears ideal. Of concern is that this is also when investigators are least likely to question the results. This counterintuitive pattern demands caution, calibration, and policy controls to prevent wrongful accusations and misuse in law enforcement. Note that the highest false positive rates are much smaller than the highest false negative rates, by a factor of 20, indicating that false positives are rarer than false negatives. 

Differential accuracy remains a serious concern, even if overall performance exceeds that of other methods. To ensure accountability, transparency is needed not only about algorithmic performance but also about how the technology is used by law enforcement. Currently, FRT is often deployed without disclosure to defendants or attorneys, undermining basic principles of due process.

Facial recognition technology is evolving rapidly, with new algorithms emerging and improving on a near-weekly basis. Future work should not only evaluate the performance of these newer models, but also develop adaptable research frameworks that can keep pace with the speed of technological change. Future work also includes testing not only the marginal effects, i.e., one single editing direction at a time, but instead combining effects. In real-world images, more than one of these features can change at once. As such, the interaction effects of various features are worth investigating. In this study, we assume the data generation and labeling algorithms have no errors. However, using a GAM to generate data could lead to having the data show a narrow type of face rather than a sample representative of the whole population. Future research could replicate our results using real data and a measure of the features. With our results, the police can create a heuristic argument about how accurate the facial recognition algorithm is with a low-quality image. Measuring the features in real images from law enforcement would allow users to know the expected accuracy at that image quality level. Finally, applying more advanced methods to evaluate FRT fairness are important direction for future research.

\section{Conclusion}

Law enforcement agencies should exercise caution when relying on FRT matches as primary evidence in criminal cases. Awareness of error rates and potential biases is crucial to prevent wrongful arrests and ensure equitable outcomes in the justice system. Future research should focus on refining facial recognition algorithms to mitigate these biases and developing clear guidelines for the appropriate use of FRT in investigative contexts. Additionally, legal and policy frameworks must be updated to reflect the risks associated with FRT misidentifications, ensuring that its use does not compromise fairness and accuracy in law enforcement.

\newpage

\appendix

\section{Factors for image degradation}\label{sec:appendix}

\subsection{Contrast and brightness}

Mathematically, an image can be represented by a rectangular matrix. For instance, an image of size $M \times N$ is a matrix with $M$ rows and $N$ columns, and the entries of the matrix are usually called pixels.
In Python's \texttt{cv2} module, contrast and brightness are controlled by two separate parameters, $\alpha > 0$ and $\beta$, respectively. Given an input image $f$, the editing on each pixel $f(i,j)$ using \texttt{convertScaleAbs} is regulated by the relationship,
\begin{equation*}
    g(i,j) = \min\{255,|\alpha f(i,j) + \beta|\},
\end{equation*}
where $g(i,j)$ denotes the $(i,j)$-th pixel of the output image. The function first applies a linear transformation to each pixel, then takes the absolute value, and finally clamps the result to the 8-bit range, i.e., all values are restricted to the range $[0,255]$. Varying $\alpha$ results in a uniform linear scaling of each pixel, which corresponds to the contrast, and varying $\beta$ results in a uniform linear translation of each pixel, which corresponds to the brightness.

Consider a $2 \times 2$ input image $f$ given by 
\scriptsize
    \begin{equation*}
        f = 
        \begin{pmatrix}
            f(1,1) & f(1,2)\\
            f(2,1) & f(2,2)
        \end{pmatrix} = 
        \begin{pmatrix}
            \min\{255,|1 \times f(1,1) + 0|\} & \min\{255,|1 \times f(1,2) + 0|\}\\
            \min\{255,|1 \times f(2,1) + 0|\} & \min\{255,|1 \times f(2,2) + 0|\}
        \end{pmatrix} = 
        \begin{pmatrix}
            12 & 24\\
            36 & 48
        \end{pmatrix}.
    \end{equation*}
    \normalsize
    In other words, an input image always corresponds to the initial set of parameters $(\alpha,\beta) = (1,0)$.
    Now, if we change the contrast parameter $\alpha$ to 2 and the brightness parameter $\beta$ to 50, the output image $g$ is given by 
    \scriptsize
    \begin{equation*}
        g = 
        \begin{pmatrix}
            g(1,1) & g(1,2)\\
            g(2,1) & g(2,2)
        \end{pmatrix} = 
        \begin{pmatrix}
            \min\{255,|2 \times f(1,1) + 50|\} & \min\{255,|2 \times f(1,2) + 50|\}\\
            \min\{255,|2 \times f(2,1) + 50|\} & \min\{255,|2 \times f(2,2) + 50|\}
        \end{pmatrix} = 
        \begin{pmatrix}
            74 & 98\\
            122 & 146
        \end{pmatrix}.
    \end{equation*}
\normalsize

To study the marginal effect of the contrast parameter, we consider the pairings 
\begin{equation} 
{\scriptstyle (\alpha,\beta) \in \{(0.25,0),(0.5,0),(0.75,0),(1,0),(1.5,0),(2,0),(2.5,0),(3,0),(3.5,0),(4,0)\}},
\end{equation}
i.e., we only vary the contrast parameter $\alpha$ but the brightness parameter $\beta = 0$ remains unchanged. Similarly, to study the marginal effect of the brightness parameter, we consider the pairings $(\alpha,\beta) \in \{(1,0),(1,25),(1,50),(1,75),(1,100)\}$, i.e., we only vary the brightness parameter $\beta$ but the contrast parameter $\alpha = 1$ remains unchanged. In particular, the pair $(\alpha,\beta) = (1,0)$ results in no editing, thus representing the baseline.

Non-linear contrast/brightness adjustments are available. For instance, \cite{Greenawalt2023} provides an approach to replicate the GNU Image Manipulation Program's non-linear contrast/brightness adjustments in Python.
For simplicity, we utilize the linear contrast/brightness adjustments provided by the \texttt{cv2} module.

\subsection{Motion blur}

While degradations in resolution come in multiple forms, we focus our discussions on motion blur because it is widely known as one of the most common blurs encountered in image processing. Contextually, police departments often perform facial recognitions based on images extracted from cameras or footage, in which motion blur is a recurrent phenomenon. 

Motion blurs are simulated on an input face by applying a 2D convolution with a specifically designed kernel used to mimic the effect of linear movement during image capture, so that the pixel values are adjusted with a pre-specified angle and blur strength $s$. In this study, we focus on horizontal motion blurs, meaning the angle is always fixed at $0^\circ$. We study the marginal effect of horizontal motion blurs by considering different blur strengths $s \in \{0,20,40,60,80,100\}$, and $s = 0$ represents the baseline as it results in no motion blurs.

For an input face, we simulate motion blurs on it by applying a 2D convolution with a kernel defined in the following way.

 Let $k$ be an $s \times s$ matrix, defined by 
 \begin{equation*}
     k_{i,j} = 
     \begin{cases}
         1, ~ &\text{if} ~ i = \dfrac{s - 1}{2},\\
         0, ~ &\text{otherwise},
     \end{cases}
 \end{equation*}
 that is, only the center row is set to ones, forming a horizontal line segment.
 The kernel is then normalized so that the sum of all its elements is 1:
 \begin{equation*}
     k_{i,j} \leftarrow \frac{k_{i,j}}{\sum_{i,j} k_{i,j}}.
 \end{equation*}
 The motion-blurred image $g(x,y)$ is produced by convolving the input image $f(x,y)$ with the kernel $k$:
 \begin{equation*}
     g(x,y) = \sum_{i = -r}^r \sum_{j = -r}^r f(x + i,y + j)k(i,j).
 \end{equation*}
 In practical terms, this means sliding the kernel over the image and, at each position, replacing the center pixel with the weighted average of its neighbors along the horizontal direction, as determined by the kernel.
 Additionally, convolution with an image often involves flipping the kernel horizontally and vertically before applying it to the image, but for symmetric kernel, such as the motion blur kernel $k$ in our case, flipping has no effect.
 The convolution operation would involve sliding the kernel over the entire image, but the range of summation indices would be relative to the center of the kernel.
 For example, for a $5 \times 5$ image with a $3 \times 3$ kernel, it looks like 
 \begin{equation*}
     g(x,y) = \sum_{i = -1}^1 \sum_{j = -1}^1 f(x + i,y + j)k(i,j).
 \end{equation*}

\subsection{Pose}
We use the ``restyle\_e4e\_ffhq'' model for pose editing. Each image is first aligned and cropped to match the encoder's training distribution. We then compute landmark-based transformations to record the geometric mapping between the original and aligned images. Next, we invert the aligned image into the StyleGAN3 latent space using the encoder, and run inference to obtain the latent code and a reconstructed image. Finally, we edit the latent code by moving it along a precomputed pose direction and synthesize the edited image. The landmark-based transformations are used to map the edited result back to the original pose and orientation, ensuring edits are consistent with the input image's geometry. During the pose editing process, some images raised exceptions and could not be processed. These problematic images were excluded from the analysis. We study the marginal effect of pose by considering different parameters $\psi = -5,-4,\ldots,4,5$, and $\psi = 0$ represents the baseline as it results in no pose editing. However, we note that since the same procedures are undergone even for $\psi = 0$, the ``baseline'' images are slightly different the original generated images. We adjust for this in our plots by adding the difference between the $\psi=0$ error rate for pose and the baseline error rate for the other image degradation factors. We add this fixed value to all the error rates and thus shift the pose curve upward for a more honest comparison. This shift w

\subsection{Resolution}

Image resolution refers to the number of distinct pixels used to represent a picture. Reducing resolution therefore discards high-frequency detail, and, even if an image is later enlarged, those lost details cannot be fully recovered. To mimic the low?resolution CCTV or social?media photographs that law?enforcement officers often submit for facial searches, we apply the following two?step process to every synthetic face:

\begin{enumerate}
    \item \textbf{Down?sample.}  Shrink the original image by a user?specified scale factor. We use Python's \texttt{cv2}  \texttt{INTER\_AREA} interpolation, which averages neighboring pixels and, therefore, best approximates a true reduction in optical resolution.
    \item \textbf{Up?sample.}  Enlarge the low?resolution image back to its original width $W$ and height $H$ with \texttt{INTER\_NEAREST} interpolation. This preserves the blocky artifacts humans associate with low?quality footage.
\end{enumerate}
We study the marginal effect of resolution by considering different scales, namely, $1\%,2\%,4\%,8\%,16\%,32\%,64\%,100\%$, and scale $= 100\%$ represents the baseline as it results in no scaling.

\newpage

\section*{Acknowledgments}

Removed for blind review.


\bibliographystyle{vancouver}
\bibliography{facialrecognition}

\begin{thebibliography}{10}

\bibitem{Garvie2022}
Garvie C.
\newblock A forensic without the science: face recognition in US criminal
  investigations.
\newblock Center on Privacy \& Technology at Georgetown Law. 2022;6.

\bibitem{clearviewaiaccuracy}
{Clearview AI}. Linden L, editor. Consecutive NIST Tests Confirm Superiority of
  Clearview AI's Facial Recognition Platform. Online; 2021.
\newblock
  \url{https://www.clearview.ai/press-room/consecutive-nist-tests-confirm-superiority-of-clearview-ais-facial-recognition-platform}.

\bibitem{nistvendortest}
Grother P. Face Recognition Vendor Test (FRVT); 2019.
\newblock
  \url{https://www.nist.gov/programs-projects/face-recognition-vendor-test-frvt}.

\bibitem{wapostinvestigation}
MacMillan D, Ovalle D, Schaffer A. Arrested by AI: Police ignore standards
  after facial recognition matches; 2025.
\newblock
  \url{https://www.washingtonpost.com/business/interactive/2025/police-artificial-intelligence-facial-recognition/?itid=sr_0_8204cd4d-7801-4d64-986f-b99edd2636a1_03abe5cd-4d5a-4f6b-91bf-6a1a2a070092}.

\bibitem{ipfrt}
Sanford A. Artificial Intelligence Is Putting Innocent People at Risk of Being
  Incarcerated; 2024.
\newblock
  \url{https://innocenceproject.org/news/artificial-intelligence-is-putting-innocent-people-at-risk-of-being-incarcerated/}.

\bibitem{Zhang_etal2018}
Zhang S, Shen X, Lin Z, M{\v{e}}ch R, Costeira JP, Moura JM.
\newblock Learning to understand image blur.
\newblock In: Proceedings of the IEEE conference on computer vision and pattern
  recognition; 2018. p. 6586-95.

\bibitem{Liu_etal2008}
Liu R, Li Z, Jia J.
\newblock Image partial blur detection and classification.
\newblock In: 2008 IEEE conference on computer vision and pattern recognition.
  IEEE; 2008. p. 1-8.

\bibitem{Punnappurath_etal2015}
Punnappurath A, Rajagopalan AN, Taheri S, Chellappa R, Seetharaman G.
\newblock Face recognition across non-uniform motion blur, illumination, and
  pose.
\newblock IEEE Transactions on image processing. 2015;24(7):2067-82.

\bibitem{Bacci_etal2024}
Bacci N, Briers N, Steyn M.
\newblock Prioritising quality: investigating the influence of image quality on
  forensic facial comparison.
\newblock International Journal of Legal Medicine. 2024:1-14.

\bibitem{terhorst2023pixel}
Terhörst P, et~al.
\newblock Pixel-Level Face Image Quality Assessment for Explainable Face
  Recognition.
\newblock ResearchGate. 2023.
\newblock Available from:
  \url{https://www.researchgate.net/publication/369678929_Pixel-Level_Face_Image_Quality_Assessment_for_Explainable_Face_Recognition}.

\bibitem{Grother_etal2019_Part3}
Grother P, Ngan M, Hanaoka K.
\newblock Face recognition vendor test (fvrt): Part 3, demographic effects.
\newblock National Institute of Standards and Technology Gaithersburg, MD;
  2019.

\bibitem{Cavazos_etal2020}
Cavazos JG, Phillips PJ, Castillo CD, O’Toole AJ.
\newblock Accuracy comparison across face recognition algorithms: Where are we
  on measuring race bias?
\newblock IEEE transactions on biometrics, behavior, and identity science.
  2020;3(1):101-11.

\bibitem{Liang_etal2023}
Liang H, Perona P, Balakrishnan G.
\newblock Benchmarking Algorithmic Bias in Face Recognition: An Experimental
  Approach Using Synthetic Faces and Human Evaluation.
\newblock In: Proceedings of the IEEE/CVF International Conference on Computer
  Vision; 2023. p. 4977-87.

\bibitem{bestrowden2018learning}
Best-Rowden L, Jain AK.
\newblock Learning Face Image Quality from Human Assessments.
\newblock IEEE Transactions on Information Forensics and Security.
  2018;13(12):3064-77.
\newblock Available from:
  \url{https://biometrics.cse.msu.edu/Publications/Face/BestRowdenJain_FaceQualityHumanAssessments_TIFS2018.pdf}.

\bibitem{howard2019effect}
Howard JJ, Sirotin YB, Vemury AR.
\newblock The effect of broad and specific demographic homogeneity on the
  imposter distributions and false match rates in face recognition algorithm
  performance.
\newblock In: 2019 ieee 10th international conference on biometrics theory,
  applications and systems (btas). IEEE; 2019. p. 1-8.

\bibitem{howard2020quantifying}
Howard JJ, Sirotin YB, Tipton JL, Vemury AR.
\newblock Quantifying the extent to which race and gender features determine
  identity in commercial face recognition algorithms.
\newblock arXiv preprint arXiv:201007979. 2020.

\bibitem{howard2022disparate}
Howard JJ, Laird EJ, Sirotin YB.
\newblock Disparate impact in facial recognition stems from the broad
  homogeneity effect: A case study and method to resolve.
\newblock In: International Conference on Pattern Recognition. Springer; 2022.
  p. 448-64.

\bibitem{howard2022evaluating}
Howard JJ, Laird EJ, Rubin RE, Sirotin YB, Tipton JL, Vemury AR.
\newblock Evaluating proposed fairness models for face recognition algorithms.
\newblock In: International Conference on Pattern Recognition. Springer; 2022.
  p. 431-47.

\bibitem{minnesotareformer}
Gross P. Reformer TM, editor; 2025.
\newblock
  \url{https://minnesotareformer.com/2025/02/04/facial-recognition-in-policing-is-getting-state-by-state-guardrails/}.

\bibitem{jackson2019challenging}
Jackson K.
\newblock Challenging facial recognition software in criminal court.
\newblock The Champion. 2019:14-26.

\bibitem{njvsarteaga}
{Superior Court of New Jersey Appellate Division}.
\newblock {State of New Jersey vs. Francisco Arteaga}.
\newblock Docket No. A-3078-21; 2023.

\bibitem{aclufacial}
ACLU. NA, editor. New Jersey Appellate Division One of First Courts in Country
  to Rule on Constitutional Rights Related to Facial Recognition Technologies.
  NA; 2023.
\newblock
  https://www.aclu-nj.org/en/press-releases/new-jersey-appellate-division-one-first-courts-country-rule-constitutional-rights.

\bibitem{haddad2020confronting}
Haddad GM.
\newblock Confronting the biased algorithm: the danger of admitting facial
  recognition technology results in the courtroom.
\newblock Vand J Ent \& Tech L. 2020;23:891.

\bibitem{celentino2015face}
Celentino JC.
\newblock Face-to-face with facial recognition evidence: Admissibility under
  the post-crawford confrontation clause.
\newblock Mich L Rev. 2015;114:1317.

\bibitem{PCAST2016}
{PCAST Working Group}. NA, editor. Forensic science in criminal courts:
  Ensuring scientific validity of feature-comparison methods. Washington, DC,
  USA: President's Council of Advisors on Science and Technology (US); 2016.

\bibitem{stern2019reliability}
Stern HS, Cuellar M, Kaye DH.
\newblock Reliability and Validity of Forensic Science Evidence.
\newblock Significance. 2019;16(2):21-4.

\bibitem{Grother_etal2019_Part1}
Grother P, Ngan M, Hanaoka K.
\newblock Ongoing face recognition vendor test (FRVT) part 1: Verification.
\newblock National Institute of Standards and Technology. 2019.

\bibitem{ulery2011}
Ulery BT, Hicklin RA, Buscaglia J, Roberts MA.
\newblock Accuracy and reliability of forensic latent fingerprint decisions.
\newblock Proceedings of the National Academy of Sciences. 2011;108(19):7733-8.
\newblock Available from:
  \url{https://www.pnas.org/doi/abs/10.1073/pnas.1018707108}.

\bibitem{amesi}
Baldwin DP, Bajic SJ, Morris M, Zamzow D. A study of false-positive and
  false-negative error rates in cartridge case comparisons. Technical report;
  2014.

\bibitem{cuellarfirearms}
Cuellar M, Vanderplas S, Luby A, Rosenblum M.
\newblock Methodological problems in every black-box study of forensic firearm
  comparisons.
\newblock Law, Probability and Risk. 2024 12;23(1):mgae015.
\newblock Available from: \url{https://doi.org/10.1093/lpr/mgae015}.

\bibitem{melzi2024frcsyn}
Melzi P, Tolosana R, Vera-Rodriguez R, Kim M, Rathgeb C, Liu X, et~al.
\newblock FRCSyn Challenge at WACV 2024: Face Recognition Challenge in the Era
  of Synthetic Data.
\newblock In: Proceedings of the IEEE/CVF Winter Conference on Applications of
  Computer Vision Workshops; 2024. Available from:
  \url{https://arxiv.org/abs/2311.10476}.

\bibitem{du2024impact}
Atzori A, Cosseddu P, Fenu G, Marras M.
\newblock The Impact of Balancing Real and Synthetic Data on Accuracy and
  Fairness in Face Recognition.
\newblock arXiv preprint arXiv:240902867. 2024.
\newblock Available from: \url{https://arxiv.org/abs/2409.02867}.

\bibitem{rahimi2024synthetic}
Rahimi P, et~al.
\newblock Synthetic to Authentic: Transferring Realism to 3D Face Renderings
  for Face Recognition.
\newblock In: Proceedings of the European Conference on Computer Vision (ECCV);
  2024. Available from: \url{https://arxiv.org/html/2407.07627v2}.

\bibitem{Karras_etal2021}
Karras T, Aittala M, Laine S, H{\"a}rk{\"o}nen E, Hellsten J, Lehtinen J,
  et~al.
\newblock Alias-free generative adversarial networks.
\newblock Advances in neural information processing systems. 2021;34:852-63.

\bibitem{Alaluf_etal2022}
Alaluf Y, Patashnik O, Wu Z, Zamir A, Shechtman E, Lischinski D, et~al.
\newblock Third time’s the charm? image and video editing with stylegan3.
\newblock In: European Conference on Computer Vision. Springer; 2022. p.
  204-20.

\bibitem{Grissom_etal2024}
Grissom~II A, Lei RF, Neto JFSR, Lin B, Trotter R.
\newblock Examining Pathological Bias in a Generative Adversarial Network
  Discriminator: A Case Study on a StyleGAN3 Model.
\newblock arXiv preprint arXiv:240209786. 2024.

\bibitem{Karras_etal2019}
Karras T, Laine S, Aila T.
\newblock A style-based generator architecture for generative adversarial
  networks.
\newblock In: Proceedings of the IEEE/CVF conference on computer vision and
  pattern recognition; 2019. p. 4401-10.

\bibitem{Mundra_etal2023}
Mundra S, Porcile GJA, Marvaniya S, Verbus JR, Farid H.
\newblock Exposing GAN-generated profile photos from compact embeddings.
\newblock In: Proceedings of the IEEE/CVF Conference on Computer Vision and
  Pattern Recognition; 2023. p. 884-92.

\bibitem{Maluleke_etal2022}
Maluleke VH, Thakkar N, Brooks T, Weber E, Darrell T, Efros AA, et~al.
\newblock Studying bias in gans through the lens of race.
\newblock In: European Conference on Computer Vision. Springer; 2022. p.
  344-60.

\bibitem{Robert_etal2023}
Robert D, Sathyamurthy S, Putha P.
\newblock MRMCsamplesize: An R Package for Estimating Sample Sizes for
  Multi-Reader Multi-Case Studies.
\newblock medRxiv. 2023:2023-09.

\bibitem{KarkkainenJoo2021}
Karkkainen K, Joo J.
\newblock Fairface: Face attribute dataset for balanced race, gender, and age
  for bias measurement and mitigation.
\newblock In: Proceedings of the IEEE/CVF winter conference on applications of
  computer vision; 2021. p. 1548-58.

\bibitem{Melzi_etal2023}
Melzi P, Rathgeb C, Tolosana R, Vera-Rodriguez R, Lawatsch D, Domin F, et~al.
\newblock Gandiffface: Controllable generation of synthetic datasets for face
  recognition with realistic variations.
\newblock In: Proceedings of the IEEE/CVF International Conference on Computer
  Vision; 2023. p. 3086-95.

\bibitem{NGA2021}
Association NG, et~al.
\newblock Age boundaries in juvenile justice systems.
\newblock Washington, DC: National Governors Association. 2021.

\bibitem{JuvenileOffendersOrdinance2022}
{Hong Kong SAR Government}. Juvenile Offenders Ordinance, Cap. 226; 2022.
\newblock Point in time: 01/07/2022.
\newblock
  \url{https://www.elegislation.gov.hk/hk/cap226?xpid=ID_1438402856544_001}.

\bibitem{GOVUK_AgeCriminalResponsibility_2025}
{UK Government}. Age of criminal responsibility; 2025.
\newblock Accessed: 2025-05-16.
\newblock Available from:
  \url{https://www.gov.uk/age-of-criminal-responsibility}.

\bibitem{SAGovMACR2025}
{Attorney-General's Department, Government of South Australia}. Minimum age of
  criminal responsibility; 2025.
\newblock Accessed: 2025-05-16.
\newblock Available from:
  \url{https://www.agd.sa.gov.au/law-and-justice/consultation/minimum-age-of-criminal-responsibility}.

\bibitem{USCensus2020}
{U S  Census Bureau}. 2020 Census Redistricting Data (Public Law 94-171)
  Summary File; 2021.
\newblock Accessed: 2025-05-07.
\newblock Available from:
  \url{https://www.census.gov/data/datasets/2020/dec/2020-redistricting-data.html}.

\bibitem{opencv_library}
Bradski G.
\newblock {The OpenCV Library}.
\newblock Dr Dobb's Journal of Software Tools. 2000.

\bibitem{lee_blur_generator}
Lee N. {Blur-Generator}: Simulates realistic blur on images using Python; 2023.
\newblock Accessed: 2025-05-07.
\newblock Available from: \url{https://github.com/NatLee/Blur-Generator}.

\bibitem{Shen_etal2020}
Shen Y, Gu J, Tang X, Zhou B.
\newblock Interpreting the latent space of gans for semantic face editing.
\newblock In: Proceedings of the IEEE/CVF conference on computer vision and
  pattern recognition; 2020. p. 9243-52.

\bibitem{SerengilOzpinar2020}
Serengil SI, Ozpinar A.
\newblock LightFace: A Hybrid Deep Face Recognition Framework.
\newblock In: 2020 Innovations in Intelligent Systems and Applications
  Conference (ASYU). IEEE; 2020. p. 23-7.
\newblock Available from: \url{https://doi.org/10.1109/ASYU50717.2020.9259802}.

\bibitem{Deng_etal2020}
Deng J, Guo J, Ververas E, Kotsia I, Zafeiriou S.
\newblock Retinaface: Single-shot multi-level face localisation in the wild.
\newblock In: Proceedings of the IEEE/CVF conference on computer vision and
  pattern recognition; 2020. p. 5203-12.

\bibitem{Deng_etal2019}
Deng J, Guo J, Xue N, Zafeiriou S.
\newblock Arcface: Additive angular margin loss for deep face recognition.
\newblock In: Proceedings of the IEEE/CVF conference on computer vision and
  pattern recognition; 2019. p. 4690-9.

\bibitem{Bolme_etal2020}
Bolme DS, Srinivas N, Brogan J, Cornett D.
\newblock Face recognition oak ridge (faro): A framework for distributed and
  scalable biometrics applications.
\newblock In: 2020 IEEE International Joint Conference on Biometrics (IJCB).
  IEEE; 2020. p. 1-8.

\bibitem{nist_frvt_visa}
of~Standards NI, Technology. FRVT 1:1 Verification Visa Results; n.d.
\newblock Accessed: 2025-05-16.
\newblock \url{https://pages.nist.gov/frvt/plots/11/visa.html}.

\bibitem{nistvendortest3}
Grother P, Ngan M, Hanaoka K. NA, editor. Face Recognition Vendor Test (FRVT),
  Part 3: Demographic Effects. NA; 2019.

\bibitem{Greenawalt2023}
Greenawalt J.
\newblock Replicating Gimp's brightness/contrast adjustment in python.
\newblock Medium. 2023 21 December.
\newblock Available from:
  \url{https://medium.com/@jsgreenawalt/replicating-gimps-brightness-contrast-adjustment-in-python-e9c962391162}.

\end{thebibliography}

\end{document}